\documentclass[conference]{IEEEtran}
\IEEEoverridecommandlockouts
\usepackage{cite}

\newcommand\red[1]{{\color{black}#1}}
\usepackage{amsmath,amssymb,amsfonts}
\usepackage{algorithmic}
\usepackage[dvipdfmx]{graphicx}
\usepackage{booktabs}
\usepackage{textcomp}
\usepackage{xcolor}
\def\BibTeX{{\rm B\kern-.05em{\sc i\kern-.025em b}\kern-.08em
    T\kern-.1667em\lower.7ex\hbox{E}\kern-.125emX}}
\usepackage{etoolbox}
\makeatletter
\patchcmd{\@makecaption}
  {\scshape}
  {}
  {}
  {}
\makeatother

\usepackage[whole]{bxcjkjatype} 
\begin{document}
\title{Is In-hospital Meta-information Useful for Abstractive Discharge Summary Generation?}

\author{
Kenichiro Ando$^{1,2}$~\;~ Mamoru Komachi$^1$~\;~ Takashi Okumura$^{3,4}$~\;~ \\ Hiromasa Horiguchi$^2$~\;~Yuji Matsumoto$^4$\\
$^1$Tokyo Metropolitan University~\;~ $^2$National Hospital Organization~\;~\\$^3$Kitami Institute of Technology~\;~$^4$RIKEN\\
{\fontsize{9.5pt}{0pt}\selectfont \texttt{\{ando-kenichiro, komachi, tokumura\}@\{ed.tmu, tmu, mail.kitami-it\}.ac.jp}}\\
{\fontsize{9.5pt}{0pt}\selectfont \texttt{horiguchi.hiromasa.nz@mail.hosp.go.jp}}\\
{\fontsize{9.5pt}{0pt}\selectfont \texttt{yuji.matsumoto@riken.jp}}
}

\maketitle
\begin{abstract}
During the patient's hospitalization, the physician must record daily observations of the patient and summarize them into a brief document called ``discharge summary'' when the patient is discharged.
Automated generation of discharge summary can greatly relieve the physicians' burden, and has been addressed recently in the research community.
Most previous studies of discharge summary generation using the sequence-to-sequence architecture focus on only inpatient notes for input.
However, electric health records (EHR) also have rich structured metadata (e.g., hospital, physician, disease, length of stay, etc.) that might be useful.
This paper investigates the effectiveness of medical meta-information for summarization tasks.
We obtain four types of meta-information from the EHR systems and encode each meta-information into a sequence-to-sequence model.
Using Japanese EHRs, meta-information encoded models increased ROUGE-1 by up to 4.45 points and BERTScore by 3.77 points over the vanilla Longformer.
Also, we found that the encoded meta-information improves the precisions of its related terms in the outputs.
Our results showed the benefit of the use of medical meta-information.
\end{abstract}

\section{Introduction}
Clinical notes are written daily by physicians from their consults and are used for their own decision-making or coordination of treatment.
They contain a large amount of important data for machine learning, such as conditions, laboratory tests, diagnoses, procedures, and treatments.
While invaluable to physicians and researchers, the paperwork is burdensome for physicians \cite{Arndt419,Ammenwerth2009}.
Discharge summaries, a subset of these, also play a crucial role in patient care, and are used to share information between hospitals and physicians (see an example in Figure \ref{table:ex_summary}).
It is created by the physician as a summary of notes during hospitalization at the time of the patient's discharge, which is known to be very time-consuming.
Researchers have begun to apply automatic summarization techniques to address this problem \cite{ando2022,Dias2020,shing2021,adams2021,moen2014,moen2016,alsentzer2018}.

Previous studies used extractive or abstractive summarization methods, but most of them focused on only progress notes for inputs.
Properly summarizing an admission of a patient is a quite complex task, and requires various meta-information such as the patient's age, gender, vital signs, laboratory values and background to specific diseases.
Therefore, discharge summary generation needs more medical meta-information, than similar but narrower tasks such as radiology report generation.
However, what kind of meta-information is important for summarization has not been investigated, even though it is critical not only for future research on medical summarization but also for the policy of data collection infrastructure.

In this paper, we first reveal the effects of meta-information on neural abstractive summarization on admissions.
Our model is based on an encoder-decoder transformer \cite{vaswani2017} with an additional feature embedding layer in the encoder (Figure \ref{fig:model_overview}).
Hospital, physician, disease, and length of stay are used as meta-information, and each feature is embedded in the vector space.
For experiments, we collect progress notes, discharge summaries and coded information from the electronic health record system, which are managed by a largest multi-hospital organization in Japan.
Our main contributions are as follows: 
%(1) We first apply the abstractive summarization method to generate Japanese discharge summaries.
%(2) We found that a transformer encoding meta-information generates higher quality summaries than the vanilla one, and clarified the benefit of using meta-information for medical summarization tasks.
%(3) We found that a model encoding disease information can produce proper disease and symptom words following the source.
%Also, we found that the model using physician and hospital information can generate symbols that are commonly written in the summary.
\begin{itemize}
\item We found that a transformer encoding meta-information generates higher quality summaries than the vanilla one, and clarified the benefit of using meta-information for medical summarization tasks.
\item We found that a model encoding disease information can produce proper disease and symptom words following the source.
In addition, we found that the model using physician and hospital information can generate symbols that are commonly written in the summary.
\item We are the first to apply the abstractive summarization method to generate Japanese discharge summaries.
 \end{itemize}

\section{Related Work}
In the studies of summarization of medical documents, it is common to retrieve key information such as disease, examination result, or medication from EHRs \cite{aramaki2009,reeve2007,gurulingappa2012,mashima2022}.
Other researchs more similar to our study targeted to help physicians get the point of medical documents quickly by generating a few key sentences \cite{liang2019,lee2018,macavaney2019,liu2018}.

Studies generating contextualized summaries can be categorized by the type of model inputs and architectures.
Some studies produced a whole discharge summary using structured data for input \cite{hunter2008,portet2009,goldstein2016}.
Other studies attempted to generate a whole discharge summary from free-form inpatient records \cite{ando2022, Dias2020,shing2021,adams2021,moen2014,moen2016,alsentzer2018}.
The free-form data is more challenging since it is noisier than structured data.
In inputting of the free-form data, extractive summarization methods, which extract sentences from the source, are commonly used \cite{ando2022,adams2021,moen2014,moen2016,alsentzer2018}.
On the other hands, an encoder-decoder model was used for abstractive summarization \cite{Dias2020, shing2021}, with a limited number of studies.
The various issues in the abstractive generation of discharge summary would be studied in the future.
%The encoder-decoder framework is expected to provide a flexible summary, but may contain some grammatical and informational errors \cite{haonan2021,dong2020,cao2020}. 

Studies using medical meta-information have long been conducted on a lot of tasks \cite{Choi2016,xu2019,scheurwegs2015,Futoma2017,zhang2018}.
In abstractive summarization on discharge summary, Diaz
et al. (2020) \cite{Dias2020} developed a model incorporating similarity of progress notes and information of the record author.
They presented an idea of integrating meta-information into the abstractive summarization model on medical documents, but did not reveal how meta-information would affect the quality of the summaries.

\section{Methods}

\begin{figure}[t]
\begin{center}
\scalebox{0.92}{
  \begin{tabular}{|p{8.8cm}|}
  \hline\vspace*{1mm}
  \#1 Bacterial meningitis\\
4/20-5/8 VCM 1250mg (q12h)\\
4/20 SBT/ABPC 1.5g single dose \\
4/20- MEPM 2g (q8h)\\
4/20-4/23 Dexate 6.6mg (q6h)\\
4/20-4/22 Nisseki polyglobin\\
4/20 1st lumbar puncture, cerebrospinal fluid glucose level 30 mg/dl (blood glucose level 95 mg/dl), cell count 2475/{\textmu}l.\\
Gram stain did not reveal any obvious bacteria, and cerebrospinal fluid culture also did not reveal any predominant bacteria.\\
The sensitivity of the gram stain for bacterial meningitis is about 60\%, and the sensitivity of the culture is not high either.\\
Also, the glucose in the cerebrospinal fluid would have been slightly lower.\\
Although no definitive diagnosis could be made, bacterial meningitis was the most suspicious disease.\\
The causative organism was assumed to be MRSA, and vancomycin and meropenem (meningitis dose) were used to cover a wide range of enteric bacteria.\vspace*{2mm}\\
  \hline
\end{tabular}
}
\end{center}
\caption{\red{Example of part of a discharge summary which is a dummy we created.}}
\label{table:ex_summary}
\end{figure}

\begin{figure}[t]
  \begin{center}
  \includegraphics[width=88mm]{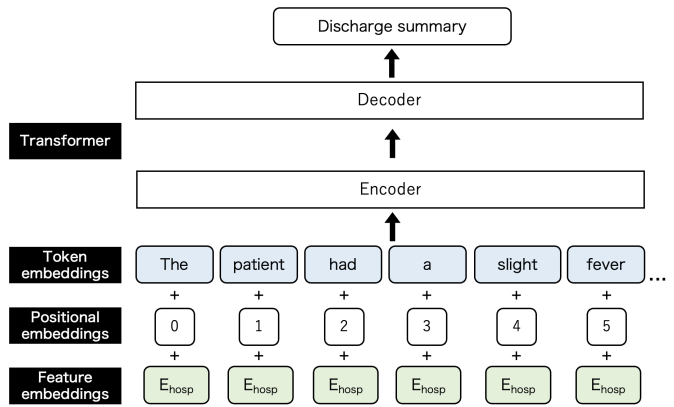}
    \caption{ Overview of our proposed method. A new feature embedding layer encoding hospital, physician, disease, and length of stay is added to the standard transformer architecture. The figure shows an example of hospital embedding.}
    \label{fig:model_overview}
  \end{center}
\end{figure}

Our method is based on the encoder-decoder transformer model.
\red{
The transformer model is known for its high performance and has been widely used in recent studies, thus it is suitable for our purpose.
}
As shown in Figure \ref{fig:model_overview}, the standard input to a transformer's encoder is created by a token sequence $T=[t_0, t_1, ..., t_i]$ and position sequence $P=[p_0, p_1, ..., p_i]$, where $i$ is the maximum input length.
The token and position sequences are converted into token embeddings $E_T$ and positional embeddings $E_P$ by looking up the vocabulary tables.
The sum of $E_T$ and $E_P$ is input into the model.

In this paper, we attempt to encode meta-information to feature embeddings.
We follow the segment embeddings of BERT \cite{devlin2019} and the language embeddings of XLM \cite{lample2019}, which provide additional information to the model. It is not a new idea but is suitable for our validation.
Our method is formulated as follows:
Let $M$ be feature type, $M \in$ \{Vanilla, Hospital, Physician, Disease, Length of stay\}, since we set five types of features.
Feature embeddings $E_M$ is created by looking up the feature table $\mathit{Table_M}=\{m_1, m_2,... , m_j, ..., | M | \}$, where $m_j$ is featue value (e.g., pysician ID, disease code, etc.) and $| M |$ is the maximum number of differences in a feature.
In our study, $| M |$ is set to four different values depending on features. 
Specifically, they are as follows.
\paragraph{Hospital}
As shown in Table \ref{table:statics}, the data includes five hospital records.
They were obtained mechanically from the EHR system.
\paragraph{Physician}
Physicians are also managed by IDs in the EHR systems.
We hashed the physician IDs into 485 groups containing 10 people each.
Specifically, as a naive strategy, we shuffled and listed the cases within each hospital, and hashed them into groups in the order of appearance of the physician IDs.
So each group has the information about the relevance of the hospitals.
The reason for employing a grouping strategy is described in Appendix A.
\paragraph{Disease}
Two types of disease information exist in our EHRs: disease names and disease codes called ICD-10\footnote{For example, botulism is A05.1 in the ICD-10 code and is connected to upper category A05, ``Other bacterial foodborne intoxications, not elsewhere classified''.}.
%Two types of disease information exist in our EHRs: disease names and disease codes called ICD-10.
We did not use any disease names in the inputs for our experiment. Instead, we encoded diseases with the first three letters of the ICD-10 code, because they represent well the higher level concept.
The initial three letters of the ICD-10 codes are arranged in the order of an alphabetic letter, a digit, and a digit, so there are a total of 2,600 ways to encode a disease.
In our data, some ICD-10 codes were missing, although all disease names were systematically obtained from the EHR system.
For such cases, we converted the disease names into ICD-10 codes using MeCab with the J-MeDic \cite{ito2018} (MANBYO 201905) dictionary.
Also, diseases can be divided into primary and secondary diseases, but we only deal with the primary diseases.
\paragraph{Length of stay}
The length of stay can be obtained mechanically from the EHR system and the maximum value was set to 1,000 days.

We set $| M |$ for vanilla, hospital, physician, disease, and length of stay to 1, 5, 485, 2,600, and 1,000, respectively\footnote{Actualy, the types of diseases and length of stay were 835 and 286, respectively. And a padding id is added.}.
The vanilla embedding is prepared for the baseline in our experiment and to equalize the total number of parameters with the other models.
The input to our model is the sum of $E_T$, $E_P$ and $E_{M}$.
We also prepare an extra model with all features for our experiments.
This takes all four feature embeddings (hospital, physician, disease, and length of stay) added to the encoder.

\section{Experimental Setup}
\subsection{Datasets and Metrics}
We evaluated our proposed method on a subset of data from National Hospital Organization (NHO), the largest multi-institutional organization in Japan.
The statistics of our data are shown in Table \ref{table:statics}\footnote{The standard deviation of the length of stay is much higher because the data set includes extremely long stays (about 26,000 days), but we found only 12 cases with length of stay above 1,000 days.}, which includes 24,630 cases collected from five hospitals.
Each case includes a discharge summary and progress notes for the days of stay.
The data are randomly split into 22,630, 1,000, and 1,000 for train, validation, and test, respectively.
Summarization performances are reported in ROUGE-1, ROUGE-2, ROUGE-L \cite{lin2004} and BERTScore \cite{Zhang2020} in terms of F1.
In addition, we also employed BLEURT \cite{sellam2020}, which models human judgment.

\begin{table}[t]
  \begin{center}
     \caption{Statistics of our data for experiment.}
    \label{table:statics}
  \scalebox{1.2}{
    \begin{tabular}{lc}
    \toprule
        Number of cases & 24,630 \\
        Average num of words in source & 1,728\\
        Average num of words in summary & 434\\
        Number of hospitals & 5 \\
        Number of physicians & 4,846 \\
        Number of diseases & 1,677 \\
        Number of primary diseases & 835\\
        Length of stay & \\
        \hspace{2mm}Average & 21\\
        \hspace{2mm}Median & 9\\
        \hspace{2mm}STD & 196\\
    \bottomrule
    \end{tabular}
    }
  \end{center}
\end{table}

\subsection{Architectures and Hyperparameters}
Due to our hardware constraints we need a model that is computationally efficient, so we employed the Longformer \cite{beltagy2020} instead of the conventional transformer.
Longformer can reduce memory usage by setting window size against calculating attention.
Our implementation of Longformer\footnote{https://github.com/ken-ando/Is-In-hospital-Meta-information-Useful-for-Abstractive-Discharge-Summary-Generation} is based on the original author's codes\footnote{https://github.com/allenai/longformer}.

In our model, number of layers, window size, dilation, input sequence length, output sequence length, batch size, learning rate and number of warmup steps are 8, 256, 1, 1024, 256, 4, 3e-5 and 1K, respectively. 
Other hyperparameters are the same as in the original Longformer, except for the maximum number of epochs is not fixed and the best epoch.
It is selected for each training using the validation data based on ROUGE-1.
Also, the original Longformer imports pretrained-BART parameters to initial values, but we do not use pre-trained Japanese BART in this study.
We used three GeForce RTX 2080 TI for our experiments.

Our vocabulary for preparing input to Longformer is taken from UTH-BERT \cite{Kawazoe2020}, which is pre-trained on the Japanese clinical records.
Since the vocabulary of UTH-BERT is trained by WordPiece \cite{wu2016}, we also tokenize our data with WordPiece.
However, the vocabulary does not include white space and line breaks, which cannot be handled, so we add those two tokens to the vocabulary, resulting in a total size of 25,002.
The vocabulary has all tokens in full characters, so we normalized full-wdith characters by converting all alphanumeric and symbolic characters to half-width for byte fallback.

\begin{table}[t]
  \begin{center}
  \caption{ Performance of summarization models with different meta-information. The best results are highlighted in bold. Each score is the average of three models with different seeds. The BS and BR indicate BERTScore and BLEURT, respectively.}
    \label{table:res_summ}
  \scalebox{1.15}{
    \begin{tabular}{lccccc}
    \toprule
         \textbf{Model} & \textbf{R-1} & \textbf{R-2} & \textbf{R-L} & \textbf{BS} & \textbf{BR}\\ 
         %Model & ROUGE-1 & ROUGE-2 & ROUGE-L & BERTScore\\ 
     \midrule   Longformer & 10.93 & 1.23 & 9.05 & 63.13 &-0.28\\
        \hspace{1mm} w/ Hospital & 13.39 & 1.41 & 10.70 & 65.19 &-0.10\\
        \hspace{1mm} w/ Physician & 14.57 & 1.02 & 10.60 & 62.30 &-0.17\\
        \hspace{1mm} w/ Disease & \textbf{15.38} & \textbf{1.96} & \textbf{12.17} &  \textbf{66.80} & \textbf{-0.07}\\
        \hspace{1mm} w/ Stay length & 14.61 & 1.25 & 10.63 & 61.94 &-0.18\\
        \hspace{1mm} w/ All features & 13.18 & 0.86 & 10.82 & 61.68 &-0.20\\
    \bottomrule
    \end{tabular}
    }
  \end{center}
\end{table}

\section{Main Results}
As shown in Table \ref{table:res_summ}, we found that all the models with encoded medical meta-information perform better in ROUGE-1, ROUGE-L and BLEURT than the vanilla Longformer.
However, in BERTScore, only hospital and disease models outperform the vanilla.
Specifically, disease information is most effective, improving ROUGE-1, ROUGE-2, ROUGE-L, BERTScore and BLEURT by 4.45, 0.73, 3.12, 3.77 and 0.21 points over the vanilla model, respectively.
This seems to be because disease information and the ICD-10 ontology efficiently cluster groups with similar representations.
In contrast, in ROUGE-2 and ROUGE-L, the model with physician embedding is inferior to the vanilla model.
This seems to be a negative effect of grouping physicians without any consideration of their relevance.
It would be better to cluster them by department, physician attributes, similarity of progress notes, etc.
Regarding low ROUGE-2 scores in all models, a previous study \cite{Dias2020} using the English data set also reported a low ROUGE-2 score of about 5\%, which may indicate an inherent difficulty in discharge summary generation.
In BERTScore, the models with the physician and the length of stay did not reach the performance of the vanilla model, suggesting that the system's outputs are semantically inferior.
The model with all features performed the lowest of all models in BERTScore.
The reason for the low score of the model with all features seems to be that its number of parameters in feature embedding was four times larger than that of the model with the individual feature, and the amount of training data was insufficient.
In BLEURT, all models with meta-information outperform vanilla, which suggests that they are more natural to humans.
%The performance of the length of stay model is high for ROUGE-1, so it is possible that the system is correctly producing more numeral terms from the date information.
%From a different perspective, the high performance of the all meta-information model suggests that there exists some non-textual bias in the collected data.
%Therefore, tasks with medical documents should be validated with test data containing a more variety of attributes more than in common domains.

\section{Precisions in Generated Words}
To analyze the influence of encoded meta-information on the outputs,
we evaluate the precisions of the generated text.
Specifically, we measure the probability that the generated words are included in the gold summary to investigate if the proper words are generated.
Some previous studies on faithfulness, which also analyze the output of summarization, have employed words or entities \cite{zhao2020,zhou2021,Goodrich2019}.
In this study, we focused on words, not entities, because we wanted to visualize expressions that are not only nouns.
%The summaries were segmented by MeCab with J-MeDic, then segmented words were grouped by MeCab for numerals and symbols, as well as by the J-MeDic dictionary for diseases and symptoms.
%The words were segmented and labeled by MeCab with the J-MeDic, and grouped by numeral, symbol, and disease and symptom labels.
The words were segmented by MeCab with the J-MeDic.
For each segmented word, the numeral and symbol labels were assigned as parts of speech by MeCab, the morphological analyzer, while the disease and symptom were assigned by the J-Medic dictionary.

\begin{figure}[t]
  \begin{center}
  \includegraphics[width=83mm]{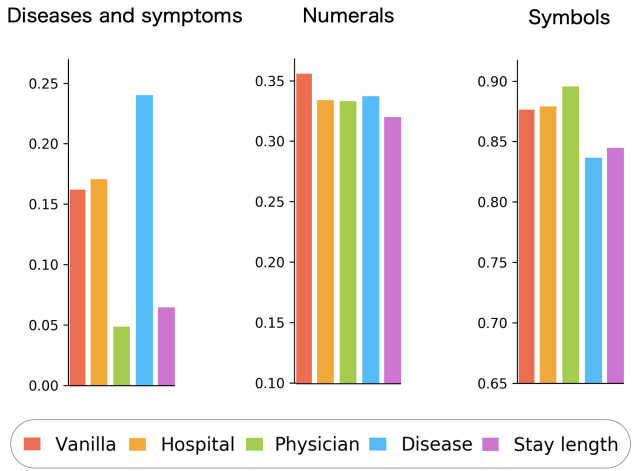}
    \caption{The precisions of words in the generated summaries. The vertical axis shows the probability that the words exist in the gold summary.}
    \label{fig:word_analysis}
  \end{center}
\end{figure}

The results, shown in Figure \ref{fig:word_analysis}, indicate that the encoded disease information leads to generate more proper disease and symptom words.
This indicates that the meta-information successfully learns disease-related expressions.
The encoded hospital or physician information also improved the precision of symbols generation.
This suggests that different hospitals and physicians have different description habits (e.g., bullet points such as ``•'', ``*'' and ``-'', punctuation such as ``。'' and ``.'', etc.), which can be grouped by meta-information.

\section{Conclusion}
In this paper, we conducted a discharge summary generation experiment by adding four types of information to Longformer and verified the impact of the meta-information.
The results showed that all four types of information exceeded the performance of the vanilla Longformer model, with the highest performance achieved by encoding disease information.
We found that meta-information is useful for abstractive summarization on discharge summaries.

Our limitations are that we used Japanese EHR, the limited number of tested features and not performing human evaluations.
As for the efficacy of the meta-information, we believe that our results are applicable to non-Japanese, but it is left as a future work.
Other meta-information may be worth verifying such as the patient's gender, age, race, religion and used EHR system, etc.
It is hard to collect a large amount of medical information and process it into meta-information, so we may need to develop a robust and flexible research infrastructure to conduct a more large scale cross-sectional study in the future.
In the discharge summary generation task, which demands a high level of expertise,
the human evaluation requires a lot of physicians' efforts and it is a very high cost which is unrealistic.
This is a general issue in tasks dealing with medical documents, and this study also could not perform human evaluations.

\section{Ethical Considerations}
On this research, informed consent and patient privacy are ensured in the following manner.
Notices about their policy and the EHR data usage are posted at the hospitals.
The patients who disagree with the policies can request opt-out and are excluded from the archive. 
In case of minors and their parents, followed the same manner.
In the case of minors and their parents are same.
To  conduct a research on the archive, researchers must  submit their research proposals to the institutional review board.
After the proposal is approved, the data is anonymized to build a dataset for  analysis.
The data is accessible only in a secured room at the NHO headquarters, and only statistics are brought out of the secured room, for protection of patients' privacy.
In the present research, the analysis was conducted under the
%IRB approval ([IRB code]) of [our affiliation].
IRB approval (IRB Approval No.: Wako3 2019-22) of the Institute of Physical and Chemical Research (RIKEN), Japan, which has a collaboration agreement with the National Hospital Organization.
This data is not publicly available due to privacy restrictions.

\bibliography{custom}
\bibliographystyle{ieee}

\newpage
\section*{Appendix}
\subsection*{A. Method of Grouping Physician IDs}
\begin{table}[t]
  \caption{Statistics on the number of cases handled by physicians. C/P denotes Cases/Physician, which indicates how many cases an individual physician has.}
    \label{table:phys}
  \begin{center}
  \scalebox{1.2}{
    \begin{tabular}{lrrrr}
    \toprule
        Hospital & Median of C/P &  Max of C/P\\
    \midrule    A & 18 & 201\\
        B & 16 & 210\\
        C & 33 & 330\\
        D & 5 & 910\\
        E & 2 & 162\\
    \bottomrule
    \end{tabular}
    }
  \end{center}
\end{table}
A most naive method of mapping physician IDs to features is without any grouping process.
The data contains 4,846 physicians, so $| M |$ was set to 4,846.
However it caused our model's training to be unstable.
This might be due to the many physician IDs appearing for the first time in the test time.
Table \ref{table:phys} shows the detailed number of cases handled by physicians.
In all hospitals, there is a large difference between the median and the maximum of cases/physician.
This indicates that a few physicians handle a large number of cases and many physicians handle fewer cases.
It is impossible to avoid physician IDs first seen at test time without some process that averages the number of cases a physician holds.
Due to this characteristic of our dataset, it was not suitable to use the physician IDs directly as features.

%\begin{table*}[t]
%  \begin{center}
%    \scalebox{1.0}{
%    \begin{tabular}{lrrrr}
%    \toprule
%        Hospital & Cases & Physicians & Median of Cases/Physician &  Max of Cases/Physician\\
%    \midrule    A & 2,015 & 40 & 18 & 201\\
%        B & 4,244 & 114 & 16 & 210\\
%        C & 2,951 & 41 & 33 & 330\\
%        D & 10,830  & 441 & 5 & 910\\
%        E & 6,390  & 672 & 2 & 162\\
%    \bottomrule
%    \end{tabular}
%    }
%  \end{center}
%   \caption{Statistics on the number of physicians and cases they handle.}
%    \label{table:phys}
%\end{table*}

\end{document}